\documentclass{article}

\usepackage{PRIMEarxiv}

\usepackage[utf8]{inputenc} 
\usepackage[T1]{fontenc}    
\usepackage{hyperref}       
\usepackage{url}            
\usepackage{booktabs}       
\usepackage{amsmath,amssymb}
\usepackage{nicefrac}       
\usepackage{microtype}      
\usepackage{fancyhdr}       
\usepackage{graphicx}       
\graphicspath{{media/}}     

\title{Mpemba Effect in Large-Language Model Training Dynamics: A Minimal Analysis of the Valley-River model}

\author{
  Sibei Liu\\
  Independent Researcher\\
  \And
  Zhijian Hu\thanks{For correspondence: \texttt{zhijianh@umich.edu}.}\\
  University of Michigan\\
}

\begin{document}
\maketitle

\begin{abstract}
Learning rate (LR) schedules in large language model (LLM) training often follow empirical templates: warm-up, constant plateau/stable phase, and decay (WSD). However, the mechanistic explanation for this strategy remains underexplored, and the choice of plateau height and decay schedule is largely heuristic. In this paper, we connect training dynamics to a thermodynamic analogy via the Mpemba effect—a phenomenon in which a hotter system cools faster than a colder one when quenched into the same bath. We analyze a class of "valley–river" loss landscapes, where sharp (valley) directions equilibrate quickly, while flatter (river) directions govern global descent. The Mpemba effect provides an explanation for the necessity of the warm-up phase and motivates a high plateau—rather than a low one—for accelerating loss decrease during decay. We show that for certain loss landscapes, there exists an optimal plateau learning rate—the "strong Mpemba point"—at which the slowest mode vanishes, resulting in faster convergence during the decay phase. We derive analytical conditions for its existence and estimate decay dynamics required to preserve the Mpemba advantage. Our minimal model and analysis offer a principled justification for plateau-based schedulers and provide guidance for tuning LR in LLMs with minimal hyperparameter sweep.
\end{abstract}


\section{Introduction}

Modern large language model (LLM) training pipelines typically rely on a three-phase learning rate (LR) schedule: warm-up, constant plateau/stable phase, and smooth decay (WSD) \cite{hu2024minicpm, wen2024understanding}. While the need for warm-up is well-understood---primarily to prevent early-stage instability and adapt to noisy gradients---the underlying reasons for adopting all three phases and how they are interconnected remain unclear. Furthermore, the choice of plateau height and duration is largely heuristic, with the plateau LR often tuned empirically, incurring significant computational cost.

Recent works have reinterpreted LLM training dynamics through the lens of effective Langevin dynamics on structured loss landscapes \cite{wen2024understanding, liu2025focus, liu2025neural}. In this framework, the loss surface consists of narrow, fast-relaxing “valleys” and broader, slow-drifting “rivers.” Learning rate plays the role of temperature, with fast directions rapidly equilibrating and slow ones drifting diffusively.

In this paper, using tools from stochastic thermodynamics, we draw an analogy between the WSD learning rate schedule and the \textbf{Mpemba effect}---a counterintuitive thermodynamic phenomenon in which a hotter system cools faster than a colder one when both are quenched into the same low-temperature bath \cite{lu2017nonequilibrium, klich2019mpemba, teza2025speedups}. This effect has been used to design optimal cooling protocols, where pre-heating paradoxically accelerates cooling, sometimes exponentially \cite{gal2020precooling, pemartin2024shortcuts}—a strategy that mirrors the WSD scheduler.

By performing a minimal “Mpemba” analysis on the valley–river model of LLM training dynamics, we identify conditions under which a high plateau LR is not only justified but necessary. We analytically derive when this phenomenon emerges from the underlying structure of the loss landscape. Moreover, our analysis predicts the existence of an \emph{optimal plateau LR}—the “strong Mpemba point”—at which the slowest dynamical mode cancels, ensuring the fastest convergence once decay begins. We also estimate possible decay dynamics, based on timescale analysis, that preserve the Mpemba advantage.

Our findings offer a theoretical justification for the constant plateau phase and provide principled guidance for selecting the plateau LR to accelerate convergence. This work establishes a minimal yet predictive framework for understanding why WSD schedules often outperform alternatives on valley–river-type loss landscapes common in LLM training.

\section{Valley--River Dynamics and Effective Quasi-1D Loss Landscape}

We consider a simplified loss landscape of the form:
\begin{equation}
L(x, y) = c(y) + \frac{1}{2}a(y)x^2,
\label{eq:loss}
\end{equation}
where \(x\) is a fast (sharp) direction with curvature \(a(y) > 0\), and \(y\) is a slow (flat) river direction. In the context of large language model (LLM) training, it has been conjectured that valley directions become sharper (i.e., higher curvature) as the loss decreases \cite{liu2025focus}. 

The stochastic gradient descent (SGD) dynamics can be approximated by a Langevin process:
\begin{align}
\dot{x} &= -\partial_x L + \sqrt{2\eta}\,\xi_x(t), \label{eq:Langevin-x}\\
\dot{y} &= -\partial_y L + \sqrt{2\eta}\,\xi_y(t),\label{eq:Langevin-y}
\end{align}
where \(\eta\) is the learning rate, acting as an effective temperature, and \(\langle \xi(t)\xi(t') \rangle = \delta(t - t')\). The drift terms are rescaled by \(\eta\) to reflect the effect of LR on search time, following the approach in \cite{yang2023stochastic}.

Although prior work has suggested that anisotropic noise in valley–river landscapes can act as a form of regularization, encouraging flatter solutions \cite{yang2023stochastic}, we ignore this effect here and assume isotropic noise for simplicity.

This loss landscape admits a timescale separation between the \(x\) and \(y\) directions, with \(\tau_x \ll \tau_y\), where \(\tau_x\) and \(\tau_y\) are the respective relaxation times to equilibrium. Integrating out the fast variable \(x\), we obtain an effective free energy landscape along the \(y\)-direction:
\begin{equation}
F_{\eta}(y) = c(y) + \frac{\eta}{2} \ln a(y).
\end{equation}

The probability density over \(y\), denoted \(p(y, t)\), evolves according to a Fokker–Planck (FP) operator associated with this effective landscape:
\begin{equation}
  \partial_t p \;=\;
  \mathcal{L}_{\eta}\,p,
  \qquad
  \mathcal{L}_{\eta}
         = \partial_y\left[\partial_y F_{\eta}(y) + \eta\,\partial_y\right].
  \label{eq:FP-operator}
\end{equation}

This Fokker–Planck operator \(\mathcal{L}_{\eta}\) governs the evolution of the probability distribution from an initial to a final state, under the influence of the learning rate (or effective temperature) \(\eta\). The approximation is valid under the assumption of timescale separation. Importantly, the effective free energy \(F_{\eta}(y)\) encodes contributions from both valley curvature \(a(y)\) and river structure \(c(y)\), serving as a minimal coupling mechanism between the two directions. While the result is quasi-one-dimensional, it captures essential two-dimensional interactions that shape the stochastic training dynamics.

\section{WSD Scheduler, Mpemba Effect, and Optimal Plateau for Stable Phase}\label{sec:MpembaMode}

Neural networks—especially deep ones like LLMs—are highly sensitive to initial updates. If the learning rate is too large at the start, gradients can explode or updates may overshoot, leading to instability or divergence. To mitigate this, the warm-up phase in the WSD scheduler is introduced to stabilize the early exploration phase of stochastic gradient descent (SGD). This has been extensively studied in recent works \cite{kosson2024analyzing}.

In the language of stochastic thermodynamics, warm-up acts like a preheating strategy. Remarkably, preheating has been shown to accelerate cooling, even exponentially, by leveraging an anomalous phenomenon known as the \textbf{Mpemba effect} or the \textbf{strong Mpemba effect}\cite{gal2020precooling, pemartin2024shortcuts}. The Mpemba effect refers to the counterintuitive observation that a hotter system can cool faster than a colder one when both are quenched into the same thermal bath.

This effect can be formulated mathematically in both discrete-state systems and continuous energy landscapes \cite{teza2025speedups}. For our purposes, we examine it through the eigenstructure of the Fokker–Planck operator \(\mathcal{L}_{\eta}\), which governs the relaxation dynamics in the river direction. Let \((u_n, \lambda_n)\) be the eigenpairs of \(\mathcal{L}_{\eta}\); then the time-dependent probability distribution evolves as
\[
  p(y,t) = \pi_{\eta}(y) + \sum_{n\ge2} a_n\,u_n(y)\,e^{-\lambda_n t}.
\]
Here, \(\lambda_2\) is the smallest nonzero eigenvalue, and the corresponding amplitude \(a_2\) and eigenfunction \(u_2\) dominate long-time relaxation.

Assume \(\lambda_2 < \lambda_3\) to ensure a clear spectral gap. Then, for long times, the distribution approximately satisfies:
\begin{equation}
p\left(y,t; \eta_0, \eta_b\right) \approx \pi_{\eta_b}(y) + a_2\,u_2\left(\eta_b\right) e^{-\lambda_2\left(\eta_b\right)t},
\end{equation}
where \(\eta_b\) is the final learning rate (analogous to the final bath temperature after decay). The eigenfunction \(u_2(\eta_b)\) and amplitude \(a_2\) determine the convergence rate to the steady-state \(\pi_{\eta_b}(y) \propto e^{-F_{\eta_b}(y)/\eta_b}\).

We define the Mpemba amplitude:
\begin{equation}
    a_2(\eta) = \int u_2(y)\,\pi_{\eta}(y)\,dy.
\end{equation}

Suppose we are choosing between two candidate plateau learning rates \(\eta_h\) and \(\eta_l\), with \(\eta_h > \eta_l > \eta_b\). If
\begin{equation}
|a_2(\eta_l)| > |a_2(\eta_h)|,
\end{equation}
then starting from a higher learning rate \(\eta_h\) leads to faster equilibration to the target distribution \(\pi_{\eta_b}(y)\). We name this \textbf{Mpemba advantage}—a regime in which a higher plateau LR yields faster convergence.

If the plateau LR is constrained to a range \([\eta_{\min}, \eta_{\max}]\), with \(\eta_{\min} > \eta_b\), then the existence of the Mpemba effect requires \textbf{non-monotonicity} of \(|a_2(\eta)|\) within that interval. Since \(|a_2(\eta_b)| = 0\), this implies that \(|a_2|\) must first increase and then decrease. A criterion is a change in the sign of the derivative of \(a_2(\eta)\), which can be expressed as:
\begin{equation}
    \boxed{
    \frac{d a_2}{d\eta} = \frac{\mathcal{K}}{\eta^2}\;\text{Cov}_{\pi_{\eta}}\left(\ln a(y), u_2(y)\right)
    }
\end{equation}
A sign flip in this covariance guarantees the non-monotonic behavior of \(|a_2|\) and is therefore a necessary condition for the emergence of the Mpemba effect. The optimal plateau LR is the value \(\eta^\star\) that minimizes \(|a_2(\eta)|\) within the allowable range:
\begin{equation}
\eta^{\star} = 
\mathop{\arg\min}_{\substack{
\eta \in [\eta_{\min},\;\eta_{\max}] \\
\eta \neq \eta_b
}} 
\left| a_2(\eta) \right|.
\end{equation}

If the minimum occurs at a value where \(|a_2(\eta)| = 0\) and \(\eta \neq \eta_b\), then the slowest mode vanishes entirely, leading to maximally fast convergence. This is the strong Mpemba effect, and the corresponding \(\eta^\star\) is called the strong Mpemba plateau. Multiple such points may exist within the LR range.

Even if \(|a_2|\) is monotonic, the minimum may still occur at \(\eta = \eta_{\min} > \eta_b\). In such cases, the selected LR is optimal within constraints but does not exhibit the Mpemba advantage.

\section{Time Scales and Estimation of LR Decaying Dynamics}

In classical Mpemba analyses—either in discrete-state systems or one-dimensional double-well landscapes (see \cite{teza2025speedups})—the pre-heating phase is typically followed by an immediate temperature drop, forming a quench that maximizes relaxation via the Mpemba effect. In our effective quasi-1D model, while equilibration along the valley (fast) direction is rapid, it is not instantaneous and we cannot ignore this finite time. This helps explain the presence of a stable phase and decay phase in the WSD scheduler: after warm-up, the distribution along the fast direction has not fully equilibrated, and a finite-duration stable phase \(t_{\text{stable}}\) at a high learning rate \(\eta\) is needed.

During the decay phase, once the valley direction is equilibrated, the learning rate must be annealed slowly enough to avoid disturbing that equilibrium—since a rapid quench could prevent the Mpemba advantage. However, annealing must not be too slow either; otherwise, each small change in \(\eta\) acts like a mini-quench, which again negates the Mpemba advantage. These considerations place upper and lower bounds on the decay time \(t_{\text{decay}}\).

The slow direction \(y\) relaxes with characteristic time \(\tau_y \approx 1/|\lambda_2| \approx 1/|\partial_y F_\eta(y)|\), while the fast direction \(x\) relaxes with \(\tau_x \approx 1/a\), governed by a simple Ornstein–Uhlenbeck process in a quadratic potential (see Eq.~\eqref{eq:Langevin-x}). To leverage the Mpemba effect, we require the following inequality to hold:
\begin{equation}
t_{\text{stable}} \gtrsim \tau_x, \quad \tau_x \lesssim t_{\text{decay}} \ll \tau_y.
\end{equation}
This ensures that, during the stable phase, the valley direction reaches quasi-equilibrium, and during the decay phase, the system remains equilibrated in \(x\) while \(y\) experiences an effective quench.

To estimate decay dynamics, we analyze the rate of change of the learning rate $\eta$. The decay must be fast enough to generate a quench for the river direction but slow enough to preserve equilibrium in the valley:
\begin{equation}
\frac{\eta(t)}{\tau_y} \ll |\dot{\eta}(t)| \lesssim \frac{\eta(t)}{\tau_x}.
\end{equation}
This yields two differential inequalities for bounding decay dynamics:
\begin{align}
\dot{\eta}(t) &= -\frac{\eta(t)}{\tau_x} = -a \eta(t), \\
\dot{\eta}(t) &= -\frac{\eta(t)}{\tau_y} \approx -k \eta^2(t),
\end{align}
where \(k = \frac{1}{2} \partial_y \ln a(y)\) (see Appendix for derivation).

Solving these equations provides bounds on the decay:
\begin{equation}
\eta^{\star} e^{-a t} \lesssim \eta(t) \ll \frac{\eta^{\star}}{1 + k \eta^{\star} t}.
\end{equation}

$\eta^{\star}$ is the optimal plateau learning rate and also serves as the initial learning rate for decay phase. Thus, to preserve the Mpemba advantage (and justify a high plateau in the stable phase), the learning rate decay must be slower than exponential decay and faster than a \(t^{-1}\) power-law decay. These bounds are not strict and should be treated as rough estimates based on the assumptions and approximations of the valley–river model.

More generally, these dynamics fall under a broader class of decay equations of the form \(\dot{\eta}(t) = -m \eta^p(t)\), with solution:
\begin{equation}
\eta(t) = \frac{\eta^{\star}}{\left[1 + (p-1) m \eta^{\star} t \right]^{\frac{1}{p-1}}}.
\end{equation}
For the fast direction, \(p = 1\) and the general solution goes back to an exponential decay form; for the slow direction, \(p = 2\). To stay within the safe decay region \(\eta(t) \gtrsim \eta^{\star} e^{-a t}\), one may either adjust the coefficient \(m\) (where \(m = a\) in this context) or use a power \(p \in [1, 2)\). We recommend decreasing \(m\) rather than increasing \(p\), as this better preserves valley equilibrium while maintaining sufficient quenching along the river direction.

In our Langevin formulation (Eq.~\eqref{eq:Langevin-x} and Eq.~\eqref{eq:Langevin-y}), we rescaled search time by multiplying it with \(\eta\) (as in \cite{yang2023stochastic}). If this rescaling is not used—as in \cite{liu2025neural}—the decay equation based on \(\tau_x\) becomes:
\begin{equation}
\eta(t) = \frac{\eta^\star}{1 + t / t_{1/2}},
\end{equation}
with \(t_{1/2} = a \eta^{\star}\). This matches the \(t^{-1}\) power-law decay recently derived by an independent research group using a different analytical approach \cite{liu2025neural}.

In conclusion, assuming the Mpemba effect exists, we make the following recommendations for the WSD scheduler during LLM training:
\[
\eta_{\text{stable}} = \eta^{\star}, \quad t_{\text{stable}} \gtrsim \frac{5}{a}, \quad \eta(t) \sim \eta^{\star} e^{-a t / 5}.
\]
The curvature \(a\) can be estimated by sampling the largest eigenvalue of the Hessian matrix during the stable phase, where the distribution is quasi-equilibrated, i.e., \(a \sim \lambda_{\max}(H)\). These conditions ensure that the fast direction \(x\) remains equilibrated and the slow direction \(y\) experiences a partial quench. This interaction between relaxation modes underpins the Mpemba effect and provides a principled guide for tuning learning rate schedules in LLM training.

\section{Discussion}

Our analysis suggests that the widely adopted WSD scheduler in LLM training can be interpreted through the lens of an anomalous thermodynamic phenomenon—the Mpemba effect. While the warm-up phase has previously been justified as a means to stabilize early-stage stochastic gradient descent \cite{kosson2024analyzing}, our framework highlights an additional role: enabling access to a high learning rate that can confer a Mpemba advantage. This aligns with the notion of optimal pre-heating strategies in thermodynamic systems exhibiting Mpemba-like behavior \cite{gal2020precooling}.

Beyond facilitating equilibration along the fast (valley) direction, the plateau phase can further enhance training by setting a learning rate that optimizes convergence in the subsequent decay phase. The Mpemba effect provides a principled criterion for selecting this plateau LR, offering guidance beyond empirical tuning. The possible existence of a strong Mpemba point—where the amplitude of the slowest relaxation mode vanishes—implies that an optimally chosen plateau LR can significantly accelerate convergence during decay.

For the decay phase, we derived approximate bounds on learning rate dynamics required to maintain the Mpemba advantage. These bounds arise from time-scale separation arguments and balance the need for preserving equilibrium in the fast direction while inducing a quench in the slow direction.

Taken together, our minimal theoretical framework unifies the warm-up, plateau, and decay phases of WSD scheduling under a single thermodynamic analogy. This provides not only explanatory insight but also a foundation for more principled scheduler design in large-scale neural network training.

\section{Caveats of Using the Mpemba Effect for Plateau Design}

While the Mpemba effect offers a compelling theoretical framework for learning rate (LR) scheduling, its application to real-world neural network training carries several caveats:

\begin{enumerate}
  \item \textbf{Model Simplification}: The valley--river model assumes a clean separation between fast (valley) and slow (river) directions with Langevin-like dynamics. Our analysis heavily depends on this separation feature. In reality, this separation may not hold, and direction coupling may violate Mpemba-related assumptions. 

  \item \textbf{Dimensionality and Complexity}: The high dimensionality of LLMs means that long-time relaxation dynamics may involve a complex mix of modes rather than a simple dominant eigenmode driven by $|a_2|$.

  \item \textbf{Inaccessibility of Slow Modes}: Identifying the slowest relaxation mode $u_2(y)$ and computing $a_2(\eta)$ is computationally challenging, particularly in online or large-scale training.

  \item \textbf{Noise and Momentum Effects}: Optimizers like Adam or momentum SGD introduce anisotropic and history-dependent noise, deviating from isotropic Langevin assumptions used in our Mpemba analysis.

  \item \textbf{Generalization vs. Optimization}: The Mpemba effect improves convergence speed to low training loss but does not directly guarantee generalization. Strong Mpemba-induced dynamics may bias training toward sharper minima.

  \item \textbf{Practical Tuning Challenges}: Even if a strong Mpemba point exists, finding it via $a_2(\eta)$ or covariance-based criteria during training remains infeasible without additional tools.

  \item \textbf{Limited Empirical Evidence}: Theoretical insights are promising, but few empirical studies have systematically tested Mpemba-based scheduling in full LLM settings.
\end{enumerate}

In summary, Mpemba-inspired plateau LR tuning offers a theoretically neat framework, but its practical implementation requires caution. Empirical validation and improved diagnostics are essential—particularly for tracking the evolution of slow modes and understanding their impact on generalization. Future work may explore extensions to adaptive optimizers, incorporation of momentum, and online estimation of key quantities such as 
$a(y),u_2(y)$ using batch-wise Hessian trace approximations. Another promising direction is to test the emergence and utility of the Mpemba effect in more general loss landscapes beyond the simplified valley–river structure considered here, including highly non-separable or rugged terrains common in deep learning. Moreover, detailed empirical studies on real-world LLM architectures are needed to determine whether strong Mpemba points can be identified in practice and whether they lead to measurable improvements in training speed or generalization. We hope this preliminary sketch of Mpemba analysis in the valley–river framework will serve as a stepping stone toward a deeper understanding of LLM training dynamics.

\section*{Acknowledgements}

N/A.

\bibliographystyle{unsrt}
\bibliography{references}

\appendix


\section*{Appendix A \,–\, Derivation of the Coarse-Grained Free Energy \texorpdfstring{$F_{\eta}(y)$}{F\_η(y)}}
\label{app:free_energy}

We provide a self-contained derivation of the effective one-dimensional
free energy landscape that arises after integrating out the fast valley coordinate in the valley–river model.

The Langevin dynamics are given by:
\begin{align}
  \dot{x} &= -\partial_x L(x, y) + \sqrt{2\eta}\,\xi_x(t),\\
  \dot{y} &= -\partial_y L(x, y) + \sqrt{2\eta}\,\xi_y(t),
\end{align}
with white noise satisfying \(\langle \xi_i(t)\xi_j(t') \rangle = \delta_{ij}\delta(t - t')\).
The potential (loss function) is defined as:
\begin{equation}
  L(x, y) = c(y) + \frac{1}{2}a(y)x^2,
  \label{eq:loss_appendix}
\end{equation}
where \(a(y) > 0\) encodes the curvature along the sharp valley direction.

Because the force is conservative and the noise is isotropic, the system satisfies detailed balance. The steady-state distribution is therefore:
\begin{equation}
  \Pi_{\eta}(x, y) = \frac{1}{Z_{\eta}}\,\exp\left[-\frac{L(x, y)}{\eta}\right],
\end{equation}
where \(Z_{\eta}\) is the normalization factor (partition function).

Since the valley direction \(x\) relaxes much faster than the river direction \(y\) (i.e., \(\tau_x = [a(y)\eta]^{-1} \ll \tau_y\)), we can treat \(y\) as quasi-static and integrate out \(x\):
\begin{align}
  \Pi_{\eta}(y)
  &= \int_{-\infty}^{\infty} \Pi_{\eta}(x, y)\,dx
   = \frac{e^{-c(y)/\eta}}{Z_{\eta}} \int_{-\infty}^{\infty}
      \exp\left[-\frac{a(y)x^2}{2\eta}\right] dx \\
  &= \frac{e^{-c(y)/\eta}}{Z_{\eta}} \sqrt{\frac{2\pi\eta}{a(y)}}.
\end{align}
We express this marginal as a Boltzmann distribution in one variable:
\(\Pi_{\eta}(y) \propto \exp[-F_{\eta}(y)/\eta]\). Taking the logarithm yields:
\begin{equation}
  F_{\eta}(y) = c(y) + \frac{\eta}{2} \ln a(y) + \text{const}.
\end{equation}
The additive constant is absorbed into \(Z_{\eta}\) and omitted.

We emphasize that this is not merely a reduction to a “1D” model. Although the effective dynamics live in one dimension after coarse-graining, the \(\ln a(y)\) term originates from integrating out the valley coordinate \(x\) and introduces a nontrivial, temperature-dependent tilt. This coupling provides a minimal connection between the valley and river directions. In other words, the fast–slow structure is what introduces the extra degree of freedom required for Mpemba-type behavior, despite the reduced dimensionality of the analysis.

The Gaussian integral contributes an entropic factor \(\sqrt{\frac{\eta}{a(y)}}\). In thermodynamic terms, we can re-express $F_{\eta}(y)$ as $F=E-TS$, where $E=c(y), T=\eta, S=(\eta/2) \ln \frac{1}{a(y)}$. The extra term \((\eta/2) \ln \frac{1}{a(y)}\) arises from the configurational entropy of the harmonic valley mode \(x\). At higher "temperature" (i.e., learning rate), this entropy contribution becomes more pronounced, tilting the effective riverbed. This entropy-driven tilt is a key mechanism enabling Mpemba-type crossovers, as discussed in Section~\ref{sec:MpembaMode} of the main text.

\section*{Appendix B \,–\, Fokker–Planck Operator for the River Coordinate}

After integrating out the fast valley coordinate \(x\), we obtain the effective one-dimensional free energy:
\[
F_{\eta}(y) = c(y) + \frac{\eta}{2} \ln a(y),
\]
where \(\eta\) acts as an effective temperature (or learning rate). On time scales \(t \gg \tau_x\), the slow coordinate \(y\) evolves according to the overdamped Langevin equation:
\begin{equation}
  \dot{y} = -\,\partial_y F_{\eta}(y) + \sqrt{2\eta}\,\xi(t),
  \qquad
  \langle \xi(t) \xi(t') \rangle = \delta(t - t').
  \label{eq:Langevin-y-effective}
\end{equation}

Let \(p(y,t)\) denote the probability density of \(y\) at time \(t\). From Eq.~\eqref{eq:Langevin-y-effective}, we obtain the Fokker–Planck equation:
\begin{equation}
  \partial_t p = \mathcal{L}_{\eta}\,p,
  \qquad
  \mathcal{L}_{\eta}
  = \partial_y \left[\partial_y F_{\eta}(y) + \eta\,\partial_y\right].
  \label{eq:FP-operator}
\end{equation}

The stationary solution is the Boltzmann distribution:
\[
\pi_{\eta}(y) = \frac{1}{Z_{\eta}} \exp\left[-\frac{F_{\eta}(y)}{\eta}\right],
\]
which satisfies \(\mathcal{L}_{\eta} \pi_{\eta} = 0\), confirming that the coarse-grained dynamics are thermodynamically consistent.

We define a self-adjoint (Hermitian) operator via a similarity transformation:
\[
\mathcal{H}_{\eta}
= -\sqrt{\pi_{\eta}}\,\mathcal{L}_{\eta}\,\frac{1}{\sqrt{\pi_{\eta}}}
= -\eta\,\partial_y^2
+ \frac{1}{4\eta} \left[F'_{\eta}(y)\right]^2
- \frac{1}{2} F''_{\eta}(y).
\]
The eigenpairs \((u_n, \lambda_n)\) of \(\mathcal{H}_{\eta}\) provide the relaxation spectrum of the system. The full solution to the time-dependent Fokker–Planck equation can thus be expressed as:
\begin{equation}
p(y, t) = \pi_{\eta}(y) + \sum_{n \ge 2} a_n\,u_n(y)\,e^{-\lambda_n t}.
\end{equation}

The smallest nonzero eigenvalue \(\lambda_2\), along with its associated amplitude \(a_2\), controls the slowest mode of relaxation and is central to the Mpemba effect analysis (see \cite{biswas2023mpemba} for detailed derivations and double-well examples).

For boundary conditions, if LLM training parameters are constrained within finite bounds, one may impose reflecting (no-flux) boundary conditions at the endpoints. These take the form:
\(J(y) = \left[\partial_y F_{\eta}(y)\,p(y,t) + \eta\,\partial_y p(y,t)\right] = 0,\)
ensuring that probability mass is conserved within the domain.

\section*{Appendix C\;–\;Derivative of the Slow–Mode Amplitude}

We analyze the derivative of the slow-mode amplitude \(a_2(\eta)\) under a quench from an initial learning rate \(\eta\) to a lower bath/final learning rate \(\eta_b\).

The stationary distribution at bath temperature \(\eta_b\) is:
\[
\pi_{\eta_b}(y) = \frac{1}{Z_{\eta_b}} \, e^{-F_{\eta_b}(y)/\eta_b}.
\]
Let \(\{u_n(y), \lambda_n\}_{n \ge 0}\) be the right eigenfunctions and eigenvalues of the Fokker–Planck operator \(\mathcal{L}_{\eta_b}\), satisfying:
$\mathcal{L}_{\eta_b} u_n = -\lambda_n u_n,$
with \(u_0 \equiv 1\), and \(0 = \lambda_0 < \lambda_1 < \lambda_2 < \cdots\). Since the eigenfunctions are defined with respect to the fixed bath temperature \(\eta_b\), they are independent of the initial temperature \(\eta\). Therefore:
\[
\frac{d}{d\eta}\,u_n(y) = 0.
\]

For \(t \ge 0\), the time-evolved probability density is:
\[
p(y, t) = \pi_{\eta_b}(y) + \sum_{n \ge 2} a_n(\eta)\,u_n(y)\,e^{-\lambda_n t},
\]
where the Mpemba amplitude for each mode is:
\(
a_n(\eta) = \int u_n(y)\,\pi_\eta(y)\,dy = \langle u_n, \pi_\eta \rangle.
\) In the following part we will derive:
\begin{equation}
\frac{d a_2}{d \eta} = \int u_2(y) \frac{d}{d\eta} \pi_\eta(y)\,dy = \int u_2(y) \left[ \frac{1}{\eta^2} \left( F_\eta(y) - \langle F_\eta \rangle_{\pi_\eta} \right) \pi_\eta(y) \right] dy.
\label{eq:derivative_a}
\end{equation}

To derive this, begin by differentiating the definition of \(\pi_\eta(y)\):
\[
\frac{d}{d\eta} \pi_\eta(y) = \frac{d}{d\eta} \left(Z_\eta^{-1} e^{-F_\eta(y)/\eta}\right).
\]

Split this into two terms:
\begin{equation}
\frac{d}{d\eta} \pi_\eta(y)
= \left( \frac{d}{d\eta} Z_\eta^{-1} \right) e^{-F_\eta/\eta}
+ Z_\eta^{-1} \frac{d}{d\eta} e^{-F_\eta/\eta}.
\end{equation}

The derivative of the normalizing constant yields:
\begin{align}
\frac{d}{d\eta} \left(Z_\eta^{-1} \right)
&= -\frac{Z'_\eta}{Z_\eta^2}
= -\frac{1}{Z_\eta} \frac{Z'_\eta}{Z_\eta}
= -\pi_\eta(y) \left\langle \frac{d}{d\eta} \ln e^{-F_\eta/\eta} \right\rangle_{\pi_\eta}.
\end{align}

Noting that \(F_\eta\) is held fixed with respect to the derivative, we compute:
\[
\frac{d}{d\eta} \left(-\frac{F_\eta}{\eta}\right) = \frac{F_\eta}{\eta^2}
\quad\Rightarrow\quad
\frac{Z'_\eta}{Z_\eta} = \left\langle \frac{F_\eta}{\eta^2} \right\rangle_{\pi_\eta}
= \frac{\langle F_\eta \rangle_{\pi_\eta}}{\eta^2}.
\]

The derivative of the exponential part is:
\[
\frac{d}{d\eta} e^{-F_\eta/\eta} = \left( \frac{F_\eta}{\eta^2} \right) e^{-F_\eta/\eta}.
\]

Combining terms, we get:
\begin{align}
\frac{d}{d\eta} \pi_\eta(y)
&= \pi_\eta(y) \left[ \frac{F_\eta(y)}{\eta^2} - \frac{\langle F_\eta \rangle_{\pi_\eta}}{\eta^2} \right] \\
&= \frac{1}{\eta^2} \left(F_\eta(y) - \langle F_\eta \rangle_{\pi_\eta} \right)\pi_\eta(y).
\end{align}

Substituting this result into Eq.~\eqref{eq:derivative_a} yields:
\begin{equation}
\frac{d a_2}{d \eta} = \frac{1}{\eta^2} \int u_2(y) \left(F_\eta(y) - \langle F_\eta \rangle_{\pi_\eta} \right) \pi_\eta(y)\,dy.
\end{equation}

Recognizing the definition of covariance, we obtain:
\begin{equation}\label{eq:C-da2}
\boxed{
\frac{d a_2}{d \eta}
= \frac{1}{\eta^2} \, \mathrm{Cov}_{\pi_\eta}(F_\eta, u_2)
= \frac{\mathcal{K}}{\eta^2} \, \mathrm{Cov}_{\pi_\eta}(\ln a(y), u_2(y))
}.
\end{equation}

Equation \eqref{eq:C-da2} is the general Mpemba-criterion formula used in Section~\ref{sec:MpembaMode} of the main text. It provides a condition for the emergence of the Mpemba effect by identifying the necessary non-monotonicity of the slow-mode amplitude with respect to the initial learning rate \(\eta\).

\section*{Appendix D\,–\,Analytical Form of Learning Rate Decay}

\subsection*{D.1 \; Two Saturated ODEs}

Assuming the fast-mode relaxation time is \(\tau_x = 1/a\), we impose the quasi-static constraint during decay:
\begin{equation}
|\dot{\eta}(t)| \lesssim \frac{\eta(t)}{\tau_x} = a \eta(t).
\end{equation}
Saturating the inequality yields the fastest decay that still maintains equilibration in the fast variable \(x\):
\begin{equation}
\dot{\eta}(t) = -a \eta(t).
\end{equation}
Solving the ODE gives an exponential decay starting from \(\eta^\star\):
\begin{equation}
\frac{d \eta}{dt} = -a \eta \quad \Longrightarrow \quad \eta(t) = \eta^\star e^{-a t}.
\end{equation}

Similarly, for the slow river direction \(y\), we require the quenching condition:
\begin{equation}
|\dot{\eta}(t)| \gg \frac{\eta(t)}{\tau_y(t)} = k \eta(t)^2,
\end{equation}
where \(k = \frac{1}{2}\,\partial_y \ln a(y)\).\footnote{To obtain this expression, note that after integrating out the fast coordinate \(x\), the effective free energy becomes \(F_\eta(y) = c(y) + \frac{\eta}{2} \ln a(y)\). Approximating near a flat region \(c'(y_0) \approx 0\), or ignoring $c'(y_0)$ and only keeping $\eta$ term, we have \(F'_\eta(y_0) \approx \frac{1}{2} \partial_y \ln a(y) \cdot \eta = k\eta\), giving \(\tau_y = 1 / F'_\eta(y) \approx 1/(k\eta)\).} To determine a safe decay rate, we again saturate the inequality:
\begin{equation}
\dot{\eta}(t) = -k \eta(t)^2.
\end{equation}
Solving this Riccati equation yields an inverse-time decay:
\begin{equation}
\frac{d\eta}{dt} = -k \eta^2 \quad \Longrightarrow \quad \eta(t) = \frac{\eta^\star}{1 + k \eta^\star t}.
\end{equation}

\subsection*{D.2 \; General Solution of Decay ODE Class}

Consider the general decay ODE:
\begin{equation}
\dot{\eta}(t) = -m \eta^p(t), \quad p > 0, \; m > 0.
\end{equation}
The integrated solutions for different values of \(p\) are summarized below:

\begin{center}
\begin{tabular}{|c|l|l|}
\hline
Exponent \(p\) & Integrated Solution (\(\eta(0) = \eta^\star\)) & Interpretation / Scheduler Type \\
\hline
\(0 < p < 1\) &
\(\eta(t) = \left[{\eta^\star}^{1-p} - m(1 - p)t\right]^{\frac{1}{1-p}}\) &
Finite-time extinction (\(\eta \to 0\) at finite \(t_{\text{stop}}\)) \\
\hline
\(p = 1\) &
\(\eta(t) = \eta^\star e^{-m t}\) &
Exponential decay \\
\hline
\(p > 1\) &
\(\eta(t) = \frac{\eta^\star}{\left[1 + (p - 1) m \eta^\star t\right]^{\frac{1}{p - 1}}}\) &
\begin{tabular}[c]{@{}l@{}}
Power-law decay \\
Special cases: \\
\quad \(p = 2\): \quad \(\eta(t) = \frac{\eta^\star}{1 + k \eta^\star t}\) (inverse-time) \\
\quad \(p = 3\): \quad \(\eta(t) = \frac{\eta^\star}{\sqrt{1 + 2k \eta^{\star2} t}}\) (inverse-square-root)
\end{tabular} \\
\hline
\end{tabular}
\end{center}

\subsection*{D.3 \; Effects of Rescaled vs. Unscaled Time}

The form of \(\eta(t)\) also depends on whether training time is rescaled by the learning rate. We compare the two cases below:

\begin{center}
\begin{tabular}{|c|c|c|}
\hline
Drift in SDE & Fast-mode relaxation time \(\tau_x\) & Fastest "Safe" \(\eta(t)\) Decay \\
\hline
No rescaling: \(\dot{x} = -\eta a x + \sqrt{2\eta} \xi\) &
\(\tau_x = \frac{1}{a\eta}\) &
\(\eta(t) = \frac{\eta^\star}{1 + t/t_{1/2}}\) (inverse-time) \\
\hline
Rescaled time: \(\dot{x} = -a x + \sqrt{2\eta} \xi\) &
\(\tau_x = \frac{1}{a}\) &
\(\eta(t) = \eta^\star e^{-a t}\) (exponential decay) \\
\hline
\end{tabular}
\end{center}

Here, \(t_{1/2} = 1/(a\eta^\star)\). The inverse-time scheduler matches the optimal decay dynamics derived via variance minimization in \cite{liu2025neural}. Both decay types are admissible if the required time-scale separation holds. While their learning curves may differ visually in \(\eta(t)\)–\(t\) plots, both preserve fast-direction equilibrium if used appropriately.

\end{document}